\documentclass[a4paper,fleqn]{cas-dc}

\usepackage[numbers]{natbib}

\usepackage{caption}
\usepackage{subcaption}
\usepackage{siunitx}
\usepackage{bm} %
\usepackage{array}
    \newcolumntype{P}[1]{>{\centering\arraybackslash}p{#1}}
    \newcolumntype{M}[1]{>{\centering\arraybackslash}m{#1}}
\def\tsc#1{\csdef{#1}{\textsc{\lowercase{#1}}\xspace}}
\tsc{WGM}
\tsc{QE}
\tsc{EP}
\tsc{PMS}
\tsc{BEC}
\tsc{DE}

\begin{document}

\let\WriteBookmarks\relax
\def\floatpagepagefraction{1}
\def\textpagefraction{.001}

\shorttitle{FaceTuneGAN: Face Autoencoder for Convolutional Expression Transfer Using Neural Generative Adversarial Networks}
\shortauthors{Olivier et~al.}

\title[mode = title]{FaceTuneGAN: Face Autoencoder for Convolutional Expression Transfer Using Neural Generative Adversarial Networks}

\author[1,2]{Nicolas Olivier}\cormark[1]
\author[1,3]{Kelian Baert}\cormark[1]
\author[1]{Fabien Danieau}
\author[2,4]{Franck Multon}
\author[1]{Quentin Avril}

\address[1]{InterDigital Inc., France}
\address[2]{Inria, Univ Rennes, CNRS, IRISA, France}
\address[3]{IMT Atlantique, France}
\address[4]{M2S, France}

\cortext[cor1,cor2]{Contributed Equally}
\cortext[cor1]{contact: nicolas.olivier@interdigital.com}

\begin{abstract}
In this paper, we present FaceTuneGAN, a new 3D face model representation decomposing and encoding separately facial identity and facial expression. We propose a first adaptation of image-to-image translation networks, that have successfully been used in the 2D domain, to 3D face geometry. Leveraging recently released large face scan databases, a neural network has been trained to decouple factors of variations with a better knowledge of the face, enabling facial expressions transfer and neutralization of expressive faces. Specifically, we design an adversarial architecture adapting the base architecture of FUNIT and using SpiralNet++ for our convolutional and sampling operations. Using two publicly available datasets (FaceScape and CoMA), FaceTuneGAN has a better identity decomposition and face neutralization than state-of-the-art techniques. It also outperforms classical deformation transfer approach by predicting blendshapes closer to ground-truth data and with less of undesired artifacts due to too different facial morphologies between source and target.
\end{abstract}


\maketitle

\section{Introduction}
\label{report-intro}

Acquiring, modeling, simulating and rendering the 3D geometry of the human face have remained key challenges in computer graphics for decades. Lately, 3D digital characters have sparked more and more interest in numerous domains, with recent research enabling higher fidelity and realism. While the modeling of actors was made available years ago~\cite{alexander2010digital}, 3D realistic faces can now be generated from scratch~\cite{li2020learning}, and thus ease the population of digital worlds. With capture technology becoming available to the mass-market, realistic digital doubles enable seamless telecommunication between individuals within virtual worlds~\cite{wei2019vr}. They also find use outside of the entertainment and telecommunication industries, e.g. sports, education, health and security. For instance, they have been used for the pre-visua-lization of plastic surgeries~\cite{bottino_new_2012}.

\begin{figure}
    \centering
    \includegraphics[width=\linewidth]{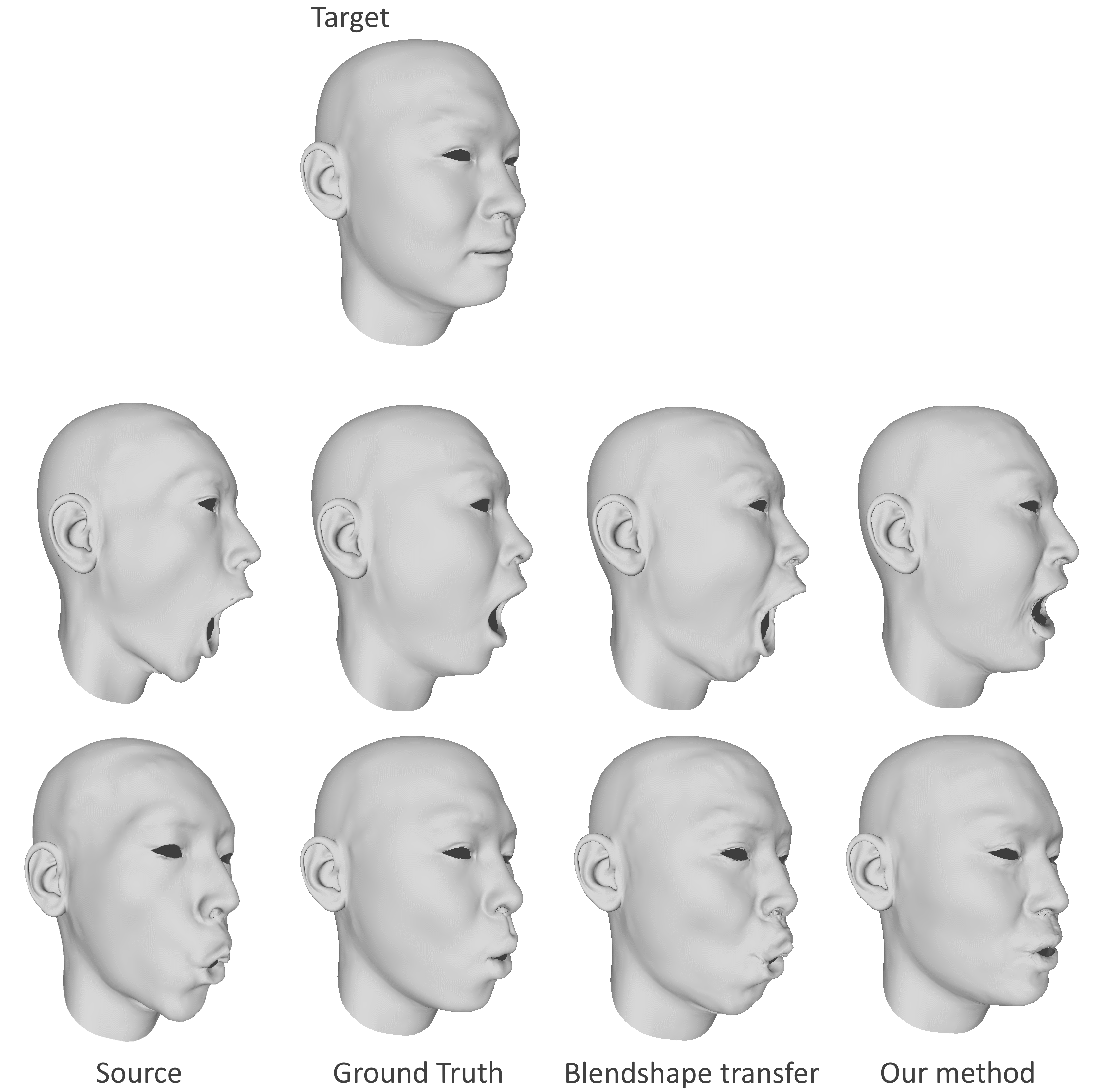}
    \caption{Blendshape transfer compared to our method. Results for the expressions {\it mouth\_stretch} (row 2) and {\it lip\_funneler} (row 3) expressions for two subjects of the FaceScape dataset are shown.}
    \label{FIG:blendshape_transfer_comparison}
\end{figure}

However, achieving a convincing representation of the 3D human face is still a difficult task. Countless attempts have resulted in appearances that, while technically impressive, fall just short of fooling the audience and end up in the so-called {\it uncanny valley}.
This is due to our considerable ability to effortlessly infer social cues and gather information when looking at human faces \cite{little_many_2011}. Variations of the 3D face shape are based on several factors, including  identity, pose, expression and age. An approach that is successfully able to decompose faces into these factors of variations would bring a better understanding of the face semantic, and eventually improve many applications such as expression transfer, expression extrapolation, performance retargeting, avatar creation \cite{danieau_automatic_2019} and personalization \cite{olivier_impact_2020}, facial recognition, or aging and de-aging. This work focuses on two factors: the identity and the expression. Our goal is to add expressions to a given 3D face model while preserving its identity and respecting its morphological constraints.

Current methods for bringing facial expressions to 3D faces often involve a lot of manual artistic work. While some high-budget productions can afford to have expressions manually sculpted by artists, this is a limiting factor for many applications and it is hardly scalable. %
Tools have been developed to adapt expressions automatically from one face to another. Blendshape transfer is a typical example~\cite{sumner2004deformation}. %
However, most of these automatic methods fail to appropriately take into account the specific morphology of the person, resulting in inaccurate depictions. %

Recently, the performance of style transfer and image-to-image translation networks has seen major performance improvements thanks to developments in deep learning. Style transfer is the task of separating a given data into a {\it content} part, and a {\it style} part, in order to stylize it by replacing its style part with another. The advent of generative adversarial networks \cite{goodfellow_generative_2014} and their many variations \cite{mirza_conditional_2014, radford_unsupervised_2016,  karras_progressive_2018} have opened the door to new architectures that outperform past approaches for style transfer in various domains \cite{isola_image--image_2018, zhu_unpaired_2020, liu_few-shot_2019}. Also, 2D generative models \cite{karras_style-based_2019, choi_stargan_2020} have come a long way in synthesizing convincing faces from compact latent representations while keeping some control on the semantics of the generated image.
In this work, we apply style transfer techniques to the 3D domain in order to decompose face geometries into structural and stylistic features. 
Adopting the language of image-to-image translation methods, we designate {\it content} as the identity dependent features of the face, and {\it style} as the elements that vary with facial expressions. 
We apply this approach to decompose identity and expression using only expression labels, allowing morphology aware expression transfer between a face with a given identity (content), and another with a given expression (style). 
We present an architecture that is able to decompose a given 3D face into decoupled content (identity) and style (expression) latent codes, and map these latent representations back into a face. Using the SpiralNet++ convolution operator \cite{gong_spiralnet_2019}, we adapt successful image-to-image translations techniques, such as the multitask adversarial discriminator introduced by Liu et al.\ \cite{liu_few-shot_2019} to the 3D domain.
Limiting ourselves to geometry (i.e.\ no texture), we achieve state-of-the-art results for decomposition and reconstruction of expressive 3D faces with a flexible approach that could be applied to other domains. This allows us to accurately transfer expressions in a morphology aware manner.

Following a review of related work in Section~\ref{relatedwork}, our approach is presented in Section~\ref{proposed-method}. Results and comparison against state-of-the-art methods are then detailed and discussed in Section~\ref{results}. Finally, limitations and conclusions are provided in Section~\ref{limitations-further-work} and~\ref{conclusion} respectively. Additional technical details are available in the appendices.

\section{Related Work}
\label{relatedwork}

Due to the prominence of face modeling as a computer graphics problem, there exists many methods for manipulating 3D geometry that are specifically tailored for the human face.
These approaches aim to map raw geometries to more semantic spaces, in order to manipulate them in a more meaningful manner than directly displacing vertices.
In this section, we first review the literature for 3D facial expression modeling to assess the limitations of current methods. We then explore the state of image-to-image translation networks. Next, we introduce common 3D mesh convolution operators used in learning-based methods. Finally, we investigate how similar approaches have been developed in the 3D domain.

\subsection{3D Face Modeling}
\label{3dfacemodeling}

Methods for modeling expressions of 3D human face have historically been largely linear or multilinear. The reader interested in this topic can refer to the survey published by Egger et al.\ \cite{egger_3d_2020}. We first consider these two categories of models, before reviewing recent nonlinear approaches.

\subsubsection{Linear and Multilinear 3D Face Models}

Blanz and Veter originally presented an approach for creating a 3D Morphable Model of the human face \cite{blanz_morphable_1999}. They used principal component analysis (PCA) to infer the distribution of facial geometry for a finite set of face scans. This method allows for manipulation of face features by manually mapping attributes to vectors in the parameter space. Rudimentary expression transfer is achieved by applying the deformation of one subject’s expressive mesh onto another. One major limitation of this first naive approach is that the expression is not adapted to the target face. For instance, when applying a mouth-opening expression from one subject to another with a larger jaw, the vertex displacements should not be the same (example on Figure~\ref{FIG:blendshape_transfer_comparison}).

To alleviate the issue of PCA components mixing identity and expression, Vlasic et al.\ extend the PCA into a multilinear model which can decorrelate shape variations caused by identity and expression \cite{vlasic_face_2005}.
Li et al.\ introduce the FLAME model, which combines a linear shape spaces, articulated parts (jaw, neck, eyeballs) and blendshapes for expressions and pose \cite{li_learning_2017}.
This allows to further disentangle identity, pose, and expression. 
Despite several improvements over time (e.g. \cite{wang_learning_2017, bolkart_robust_2016}), these approaches fall short when it comes to modelling subtle facial details. Because variations of the face shape in the real world are nonlinear in nature, we now look at architectures that introduce nonlinearities for more accurate representations of the face shape.

\subsubsection{Nonlinear 3D Face Models}

In recent years, nonlinear approaches to modeling the human face have outperformed previous methods. In 2017, Li et al.\ present the linear FLAME model. Since then, new models that make use of recent developments in deep learning have outperformed FLAME. Ranjan et al.\ manage to represent 3D faces with 75\% fewer model parameters than previous attempts, and outperform previous approaches for reconstruction and interpolation of expressions, in addition to demonstrating the ability to synthesize new faces \cite{ranjan_generating_2018}. Adopting a multi-scale autoencoder approach, they use graph convolutions on face meshes and introduce new efficient down-sampling and up-sampling operators.
Abrevaya et al.\ use an Auxiliary Classifier GAN \cite{odena_conditional_2017} to model non-linear variations of 3D face geometry while decoupling identity and expression factors \cite{fernandez_abrevaya_decoupled_2019}. Given an identity vector, an expression vector and noise, the generator outputs 3D coordinates of the mesh. These coordinates are mapped to a two-dimensional image, which is fed to a discriminator that classifies it into an identity class, an expression class and real/fake classes using standard 2D Convolutional Neural Networks (CNN) techniques. Although these models achieve state-of-the-art reconstruction results and a decoupled face representation, they require optimizing an input noise vector to fit a given face. Jiang et al.\ address this by adopting an autoencoder architecture, similar to CoMA’s \cite{ranjan_generating_2018}, with the added ability to decompose the input face into identity and expression latent representations \cite{jiang_disentangled_2019}. A fusion module then reconstructs these representations into a face. Zhang et al.\ extend this decomposition approach with an architecture that enforces distributional independence between identity and expression attributes by design \cite{zhang_learning_2020}. They obtain great performance on the task of neutralizing the expression or identity of a given mesh. Their method consists in extracting both an identity mesh and an expression mesh from the given face, and adding them together for reconstruction. However, the decoded identity and expression mesh are respectively expression-agnostic and identity-agnostic by design. Thus, their sum cannot accurately capture the coupling between the two factors. This limits the performance of expression transfer for expressions that differ substantially from the neutral expression. 

\subsection{Style Transfer and Image-to-Image Translation}
\label{2d_style_transfer}

Style transfer methods traditionally transform an input image so that it preserves its content while adopting stylistic features of another input \cite{gatys_image_2016}.
Image-to-image translation networks are a generalization of these procedures. Such networks have been able to successfully map between the specific local attributes of multiple sets of images \cite{zhu_unpaired_2020, liu_few-shot_2019}. In other words, they can transfer an image from one class to another while keeping the content intact. Typical examples include translating between animal classes, changing the season of a photographed landscape or altering specific features of a human face (e.g. hair color, morphology, age). Isola et al.\ released the \emph{pix2pix} framework, outperforming previous methods \cite{li_precomputed_2016,  ledig_photo-realistic_2017, zhu_generative_2018} by applying Conditional GANs to the image translation problem \cite{isola_image--image_2018}. Furthermore, they have demonstrated outstanding performance in a large variety of domains, even compared to domain-specific methods. Unsupervised image-to-image translation methods have obtained competitive results without requiring pair supervision \cite{taigman_unsupervised_2016, liu_unsupervised_2018, liu_few-shot_2019, zhu_unpaired_2020}. This opens the door to wider applications, since less time needs to be spent manually annotating the data. These methods use specific constraints to preserve features of the input data in the absence of a target ground truth while training. If we consider a face's identity to be the content, and its expression to be a style, they appear to be good candidates to solve our problem. They are although limited to 2D domains.

\subsection{Mesh Convolution Operators}
\label{meshconvops}

Adapting successful image-to-image translation techniques to 3D geometry could enable better accuracy in transferring attributes from one shape to another. To tackle this ill-posed problem, one of the challenges is to extract style features from the shape structure. Analogous to the convolutional layers that are used to extract style information from images, several approaches have been developed for performing convolutions on 3D meshes over the past few years. Many papers apply spectral graph convolutions \cite{bruna_spectral_2014} to 3D faces \cite{jiang_disentangled_2019, zhang_learning_2020, ranjan_generating_2018}. Others map geometries to the 2D domain and apply standard 2D convolution operators \cite{sinha_deep_2016, fernandez_abrevaya_decoupled_2019, moschoglou_3dfacegan_2019}.

In 2018, Lim et al.\ introduced SpiralNet, a new convolution operator specialized for 3D meshes \cite{lim_simple_2018-1}. Gong et al.\ later released SpiralNet++ to refine the approach \cite{gong_spiralnet_2019}. This new operator captures local geometric features by using pre-computed spiral sequences on the mesh surface. Their results are competitive with previous state-of-the-art methods for reconstruction on 3D face datasets. With the same number of parameters, they outperform previous methods while running several times faster.

\subsection{3D Style Transfer}
\label{3d_style_transfer}

While there is a growth in approaches to apply convolutions to the 3D domain, there currently exist few attempts at 3D shape-to-shape style transfer.
Recently, Segu et al.\ developed a 3D style transfer architecture that uses a PointNet \cite{qi_pointnet_2017} encoder and a decoder with Adaptive Instance normalization (AdaNorm) \cite{xu_understanding_2019} to perform shape translation on static objects \cite{segu_3dsnet_2021}. In 2019, Moschoglou et al.\ introduced 3DFaceGAN for representation, generation and translation of 3D facial surfaces allowing to transfer expression and resolution levels between facial scans.\cite{moschoglou_3dfacegan_2019}.
While they obtained state-of-the-art results at the time for the task of reconstruction, the expression transfer capabilities are limited. Their approach to multi-label face translation considers mapping neutral faces to a limited amount of discrete expressions. However, the identity and expression factors are not explicitly decoupled, as they share the same latent space.

Our aim is to create a model that can take arbitrary faces as input, and can disentangle identity and expression to perform expression transfer. This analysis of the literature suggests that techniques from recent image-to-image translation networks could be adapted to the 3D domain. 

\section{Proposed Method}
\label{proposed-method}
We propose an architecture that adapts successful image-to-image translation techniques to 3D geometry in order to map local features of a 3D face shape (content) from one facial expression to another (style). Specifically, we adapt the base architecture of FUNIT \cite{liu_few-shot_2019}, using SpiralNet++ \cite{gong_spiralnet_2019} for our convolution and sampling operators. First, we define several terms and notations used in our approach. We then present our architecture and the objectives that are optimized during training.

\subsection{Notations and Definitions}

In our approach, we adopt language analogous to style transfer and image-to-image translations papers. We designate the {\it content} of the faces as their identity: these are the structural features of the face that do not depend on the facial expressions and should be preserved when performing style transfer. The term {\it style} corresponds to the features that vary with facial expressions.

Let $\chi$ denote a domain of 3D shape, defined as triangular meshes. All shapes within these domains have previously been fitted to the same topology. Thus, any sample $x \in \chi$ is represented by a matrix of 3D coordinates of shape $(V, 3)$, $V$ being the number of vertices in our topology.

We propose to learn encoding functions $E_c$ and $E_s$ that map a sample $x \in \chi$ to its respective content and style latent codes $E_c(x)$ and $E_s(x)$. We also learn a corresponding decoding function $Dec$ that reconstructs a 3D geometry from the latent codes. These functions should ideally satisfy the reconstruction constraint $Dec(E_c(x), E_s(x)) = x$, meaning that we are able to encode and decode our geometry in a lossless manner. To achieve expression transfer, we use the common method of encoding a source mesh, swapping its style code with one that corresponds to a different expression, then decode to reconstruct a mesh.

\textbf{Average vertex distance}: we use average vertex distance to measure distances between two geometries. Given two triangle meshes $x$ and $y$ with $V$ vertices under the same topology, the average vertex distance is defined as:
\begin{equation}
    \text{AVD}(x, y) = \frac{1}{V} \sum_{i=1}^{V} ||x_i - y_i||_2
    \label{eq:avd}
\end{equation}
where $x_i$ and $y_i$ respectively denote the $i$th vertex of $x$ and $y$ and $||\cdot||_2$ is the L2 Euclidean distance.

\subsection{Architecture}

\begin{figure}
    \centering
    \includegraphics[width=\linewidth]{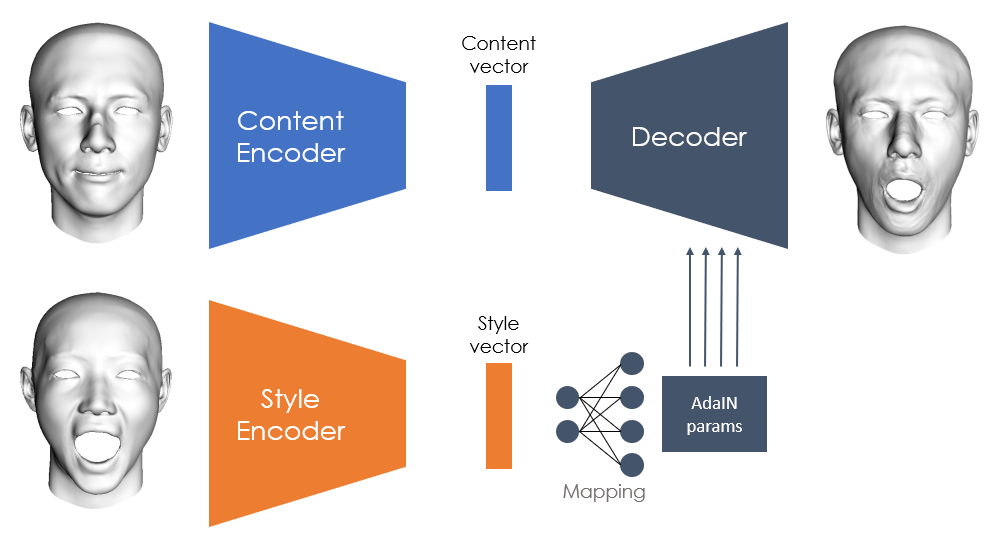}
    \caption{Our network architecture. The encoders extract and compress the features of their input into low-dimensional content and style vectors. A decoder reconstructs a mesh from these compact representations. The style information is passed to the decoder through AdaIN normalization layers. 
    More details of the architecture are given in the Appendix (Section \ref{appendix:archi-details}).}
    \label{FIG:archi}
\end{figure}

\begin{figure}
    \centering
    \begin{subfigure}{.23\textwidth}
        \centering
        \includegraphics[width=\linewidth]{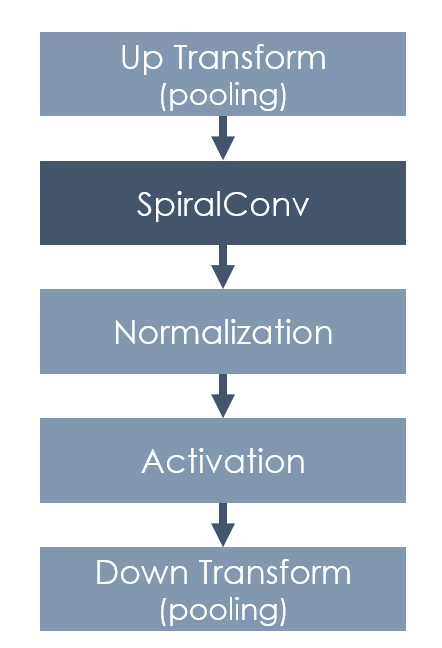}
        \caption{SpiralBlock (the layers depicted in light blue are all optional)}
        \label{FIG:spiralblock}
    \end{subfigure}
    \begin{subfigure}{.23\textwidth}
        \centering
        \includegraphics[width=\linewidth]{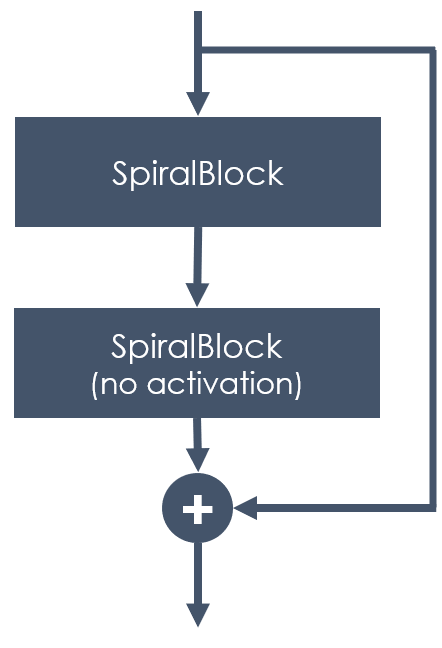}
        \caption{SpiralResBlock}
        \label{FIG:spiralresblock}
    \end{subfigure}
    \caption{The building blocks of the network.}
\end{figure}

We introduce an autoencoder that is able to separate the content (identity) and style (expression) features of the input facial shapes. The complete architecture is presented in Figure \ref{FIG:archi}. The specific composition of the modules is given in the appendix (Section~\ref{appendix:archi-details}, Figures \ref{FIG:archi_specifics_facescape} and \ref{FIG:archi_specifics_coma}), with tweaks depending on the dataset used for training. We adopt SpiralNet++ \cite{gong_spiralnet_2019} for all of our convolutions and pooling \cite{garland1997} layers and use SpiralBlock
as the main building block of our mapping functions. A SpiralBlock (Figure \ref{FIG:spiralblock}) is composed of an optional up-sampling pooling layer, followed by a convolutional layer, normalization, activation and an optional down-sampling layer. Additionally, we use the SpiralResBlock variant described in Figure \ref{FIG:spiralresblock}. 

\subsubsection{SpiralNet-based Autoencoder for 3D Faces}
The autoencoder is composed of 3 parts: The content encoder, that takes a face as input, and extract its content (identity), the style encoder, that takes another face and extract its style (expression), and finally the decoder, which generates a face from the outputs of the content and style encoders (see Figure~\ref{FIG:archi}).

\textbf{Content Encoder}: the content encoder is composed of several downscaling SpiralBlocks (the exact number depends on the dataset, see appendix~\ref{appendix:archi-details}), two SpiralResBlocks and a Multi-Layer Perceptron (MLP). The SpiralBlocks capture local information in the geometry. We then flatten the spatial and feature dimensions together and use the MLP to compress the information into the content representation.

\textbf{Style Encoder}: similarly, the style encoder is made up of a series of downscaling SpiralBlocks. However, as style information must be global in nature, instead of flattening the spatial and feature dimensions together, we compute the mean along the spatial dimension. Similarly to the content encoder, the resulting vector is then fed to a MLP in order to control the size of the latent space.

\textbf{Decoder}: the decoder upscales the content code into a 3D geometry using SpiralBlocks that are conditioned by the style code using Adaptive Instance normalization (AdaIN) \cite{huang_arbitrary_2017}. Given a sample $x$ that is passing through the network, AdaIN first normalizes the activations in each channel of $x$ to a zero mean and unit variance. The activations are then scaled on a per-channel basis. We use a mapping function $M$ that maps a style code $y$ into $(\mu,\sigma)$ parameters for every channel of each AdaIN layer. Hence the following equation:
\begin{equation}
    \text{AdaIN}(x,y) = M_{\sigma}(y)\frac{x - \mu(x)}{\sigma(x)} + M_{\mu}(y)
    \label{eq:ada_in}
\end{equation}
$M$ is a learned affine function composed of multiple fully connected layers, taking the style latent code as input. Since the AdaIN transformation operates on whole channels, the style code alters global appearance information while the local features (e.g. the shape of the chin) are determined by the content code.
The mapping function and decoder are not conditioned by a discrete class label. As a result, style codes are not domain-specific, in contrast with several existing methods for image translation \cite{hui_unsupervised_2018}, image generation \cite{choi_stargan_2020} and shape transfer \cite{segu_3dsnet_2021}. Having a single style space allows us to interpolate between style codes.

Together, the content encoder, style encoder, style mapping and decoder make up the generator $G$ of our adversarial network.

\subsubsection{Discriminator}

Similar to FUNIT \cite{liu_few-shot_2019}, we implement a multi-task adversarial discriminator $D$. Its role is to both enforce that the output mesh belongs to the distribution of the target style class, and that its geometry cannot be distinguished from a real scan. $D$ solves as many binary classification tasks as there are style classes in the dataset. For each of these style classes, $D$ outputs a classification of whether the geometry is a real sample of that class, or a translation output from the generator.

Let $s$ denote a style class.
\begin{itemize}
\item When updating the discriminator with a translation output of class $s$ from the generator, we penalize $D$ if and only if its $s$-th output is positive and ignore the predictions for other classes. Given a real geometry of style class $s$, we penalize it if its $s$-th output is negative. This way, the discriminator learns to distinguish real from generated meshes.
\item When updating the generator, we perform a style translation using a sample of style class $s$. We then penalize $G$ if the $s$-th prediction output from $D$ is negative. This encourages the generator to output realistic meshes.
\end{itemize}

Similarly to our encoders, the discriminator is composed of SpiralBlocks that gradually downscale its input mesh in the vertex dimension while adding features, followed by a linear layer for classification. More details can be found in Appendix \ref{appendix:archi-details}. 

\subsection{Loss Functions}

In this section, we define the loss functions that are used for training our network. 
Let $x$ and $s$ denote two samples respectively taken from our content and style sets. Let $x_r$ denote the reconstructed mesh obtained by encoding and decoding $x$:
\begin{equation}
    x_r = Dec(E_c(x), E_s(x))
    \label{eq:def_xr}
\end{equation}
We also define $x_t$, the mesh obtained after translating $x$ into the style class of $s$:
\begin{equation}
    x_t = Dec(E_c(x), E_s(s))
    \label{eq:def_xt}
\end{equation}

\textbf{Reconstruction loss}: we define the reconstruction loss for our generator as the following:
\begin{equation}
    L_{\text{rec}}(G) = ||x_r - x||_2
    \label{eq:loss_rec}
\end{equation}
By minimizing this loss, we force the encoders and decoder to extract relevant information in order to compress to and from our latent spaces with the least possible loss.

\textbf{Cycle consistency loss}: we adopt the cycle consistency loss \cite{zhou_learning_2016} \cite{zhu_unpaired_2020}, defined as:
\begin{equation}
    L_{\text{cycle}}(G) = ||Dec(E_c(x_t), E_s(x)) - x||_2
    \label{eq:loss_cycle}
\end{equation}
Its purpose is to ensure that the generator is able to translate $x$ to the style class of $s$ and back to its original style class with minimal content information loss.

\textbf{Style reconstruction loss}: we introduce a novel style reconstruction loss to encourage $G$ to preserve the specific style features of the input style mesh in the style latent space: 
\begin{equation}
    L_{\text{srec}}(G) = ||E_s(x_t) - E_s(x)||_1
    \label{eq:loss_srec}
\end{equation}

\textbf{Adversarial loss}: our adversarial loss is a conditional loss given by:
\begin{equation}
    L_{\text{adv}}(G, D) = \text{E}\left[\; -\log(D_s(s)) \;\right] + \text{E}\left[\; \log(1-D_s(x_t)) \;\right]
    \label{eq:loss_adv}
\end{equation}
where $D_s(\cdot)$ denotes the discriminator output for the style class of $s$ and $\text{E}[\cdot]$ the mean over the current batch.

\textbf{Feature matching loss}: we use a feature matching loss that leverages our multi-task adversarial discriminator to encourage our generator to output meshes that belong to the correct style class:
\begin{equation}
\begin{split}
    L_{\text{feat}}(G) = 
    & \text{E}\left[\; ||D^f(x_r) - D^f(x)||_1 \;\right] \\
    & + \text{E}\left[\; ||D^f(x_t) - D^f(s)||_1 \;\right]
\end{split}
    \label{eq:loss_feat}
\end{equation}
where $D^f(\cdot)$ denotes the last feature layer of the discriminator, prior to classification. By minimizing this loss, we enforce that the generator preserves style features when reconstructing the input and includes style features of the target when translating.

\textbf{Discriminator regularization loss}: we regularize the training by adopting the gradient penalty regularization loss $L_{\text{reg}}$ introduced by Mescheder et al.\ \cite{mescheder_which_2018} and used in FUNIT~\cite{liu_few-shot_2019}.

\textbf{Laplacian Smoothing loss}: finally, we use a Laplacian Smoothing loss on the translated mesh \cite{desbrun_implicit_1999} \cite{nealen_laplacian_2006}. Specifically, we use Pytorch3D’s implementation of uniform mesh Laplacian Smoothing, which consists in minimizing the distance between every vertex and the centroid of its neighbors \cite{ravi_accelerating_2020}. Empirically, we have determined that this constraint makes the generator perform better at reconstruction at the cost of some high frequency resolution.

\textbf{Full objective}: our network is trained by solving the following minimax optimization problem:
\begin{equation}
\begin{split}
    \min_{D}\max_G
    L_{\text{reg}}(D)
    + \lambda_{\text{adv}} L_{\text{adv}}(G, D)
    + \lambda_{\text{feat}} L_{\text{feat}}(G) \\
    + \lambda_{\text{srec}} L_{\text{srec}}(G)
    + L_{\text{rec}}(G)
    + L_{\text{cycle}}(G)
\end{split}
    \label{eq:minmax}
\end{equation}
where $\lambda_{\text{adv}}$, $\lambda_{\text{feat}}$ and $\lambda_{\text{srec}}$, are hyper-parameters. The discriminator is trained for one iteration every time we train the generator.

\section{Results}
\label{results}

This section reports the evaluation of our proposed method. We first describe the datasets that were used. Then, we give details for the implementation with which the evaluation was conducted. Finally, our results are evaluated in a series of experiments and compared to existing methods. 

\subsection{Datasets}

We evaluate the capabilities of our approach on two publicly available datasets:

\textbf{FaceScape} \cite{yang_facescape_2020}: this dataset includes 16,940 topologically uniformed 3D face models, captured from 847 subjects performing 20 facial expressions. Displacement and texture maps are also available, though they are not used in this work. Out of the 847 subjects, we discard 40 due to issues with some of the scans. We select 10\% of the remaining scans as our test set.

\textbf{CoMA} \cite{ranjan_generating_2018}: this dataset contains dynamic sequences of 12 subjects, each performing 12 facial expressions. In total, it comprises 144 sequences which add up to more than 20k face scans. While the small number of subjects does not allow for great generalization of identity features, the facial expressions are more extreme and asymmetrical than those of FaceScape. In our case, we need discrete expression labels for training. First, we select the first frame of each sequence as samples of neutral expression for the subject. Then, on each sequence, we select the frame with the largest average vertex distance to the first (neutral) frame. Since it is not always the best match to represent the expression, we manually verify all sequences and adjust this selection. We also split up the {\it mouth\_up}, {\it mouth\_middle} and {\it mouth\_down} expressions into their left and right variants, adding up to a total of 17 expressions, though some do not exist for all subjects (see Figure \ref{FIG:coma}). Finally, we sample 10 frames before and after the selected one in each sequence to add diversity and noise. We obtain a total of 3,
720 scans, which we randomly split into train and test sets by a 9:1 ratio.

For fair comparison, we limit the dimensionality of our latent space.  On the CoMA dataset, both our content and style spaces have 4 dimensions, adding up to a total dimensionality of 8. On the FaceScape dataset, Kacem et al.\ \cite{kacem_disentangled_2021}, use a single latent space of 25 dimensions. However, this latent space is only used to represent neutral faces. Nonetheless, we limit ourselves to 20 content dimensions and 5 style dimensions for these comparisons.

\subsection{Implementation Details}

We set the weights in Equation \eqref{eq:minmax} to $\lambda_{\text{adv}} = 1.0$, $\lambda_{\text{feat}} = 1.0$, $\lambda_{\text{srec}} = 0.4$. All of our spiral convolutions use a sequence length of 9 with no dilation (see \cite{gong_spiralnet_2019} for more information). We train our generator and discriminator using ADAM optimizers \cite{kingma_adam_2015}, with a learning rate of $\num{1e-4}$ and a weight decay of $\num{5e-5}$. All weights are initialized using the Kaiming method \cite{he_delving_2015}. We train with a batch size of 8, until no significant improvements are seen on our reconstruction, neutralization and style transfer metrics. The training duration depends on the dataset. On FaceScape, we train for 70 epochs over a duration of approximately 36 hours. On CoMA, the network is trained for 480 epochs over 8 hours. Fine-tuning was done on a NVIDIA GeForce RTX 2080 Ti, a Tesla P100 GPU and a Tesla V100. The final training runs were done on the 2080 Ti for CoMA, Tesla V100 for FaceScape.

\subsection{Experiments}

First, we assess our autoencoder’s ability to reconstruct an input mesh and compare it with state-of-the-art methods. Second, we conduct the expression neutralization task and compare our results with several baselines. Third, we evaluate our performance on the more general expression transfer task, and we compare the results to a blendshape transfer method.

\subsubsection{Reconstruction}

In order to ensure that our autoencoder is able to compress the information with as little information loss as possible, we calculate the reconstruction error as follows:
\begin{equation}
    E_\text{rec} = \frac{1}{|S|}\sum_{x \in S} \text{AVD} (x, Dec(E_c(x), E_s(x)))
    \label{eq:def_error_rec}
\end{equation}
where $S$ is a testing set of face geometries.

\begin{figure}
    \centering
    \includegraphics[width=\linewidth]{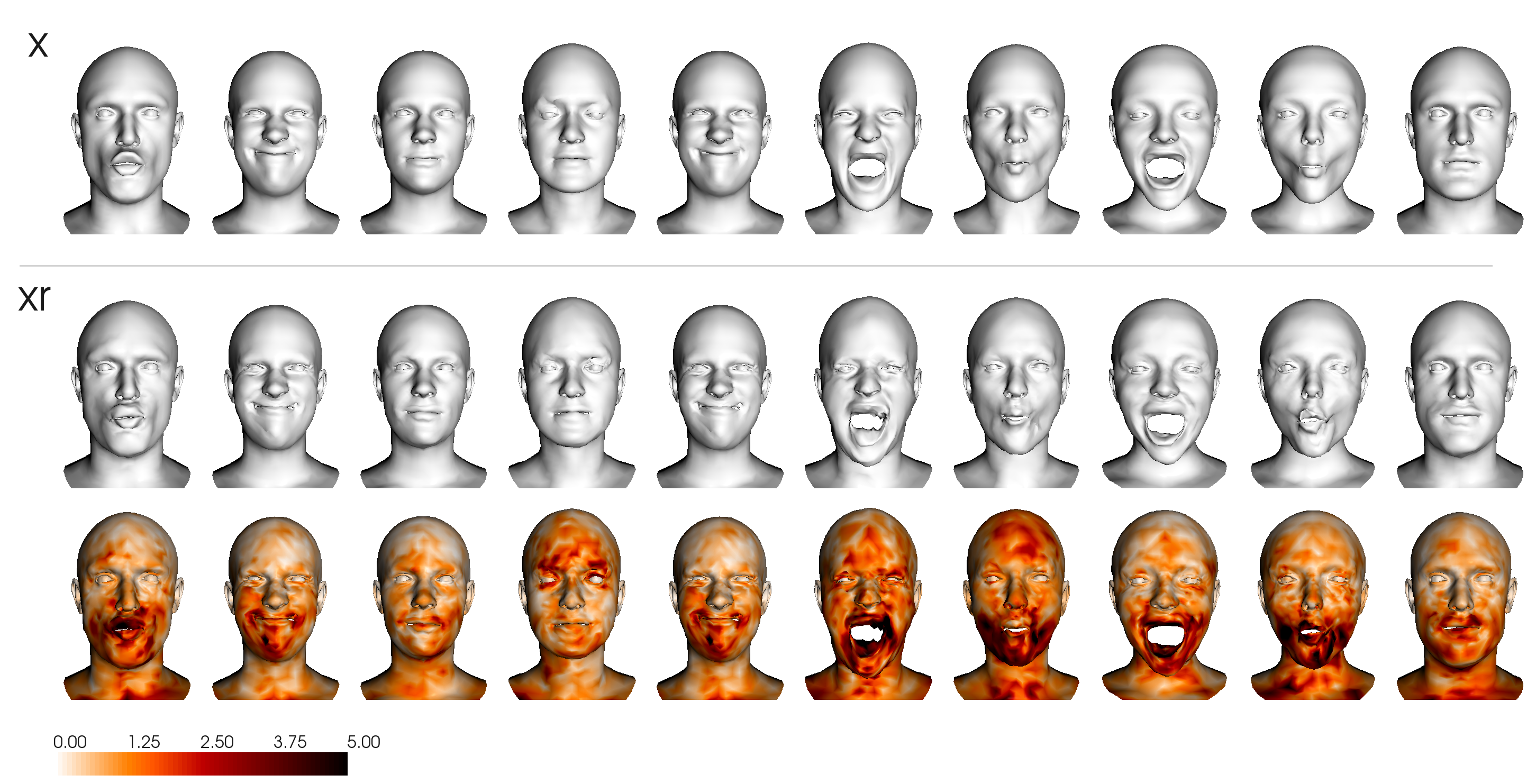}
    \caption{Reconstruction examples on CoMA. The first row shows the input mesh. The reconstruction is displayed in the second row, while the third row shows an error map relative to the input. Values are in millimeters.}
    \label{FIG:reconstruction_coma}
\end{figure}

Figure \ref{FIG:reconstruction_coma} shows error maps of reconstruction samples on the CoMA dataset. In Table \ref{TAB:reconstruction_coma}, we list quantitative results compared with other methods for both reconstruction and disentanglement of identity and expression. All methods use a total latent space size of 8. On CoMA, we attain reconstruction results in the range of other disentanglement methods: we perform better than FLAME~\cite{li_learning_2017} and Jiang et al.~\cite{jiang_disentangled_2019} but worse than Zhang et al.~\cite{zhang_learning_2020}. Note that the latter benefits from being able to use the entire CoMA database, while we only select a few frames per sequence to fit  discrete expression labels. We also do not require each mesh to be paired with its neutral ground truth for training. For comparison, we include the results of the original SpiralNet++~\cite{gong_spiralnet_2019} architecture, which obtains better performance on the reconstruction but does not disentangle identity and expression.

\begin{figure*}
    \centering
    \includegraphics[width=\linewidth]{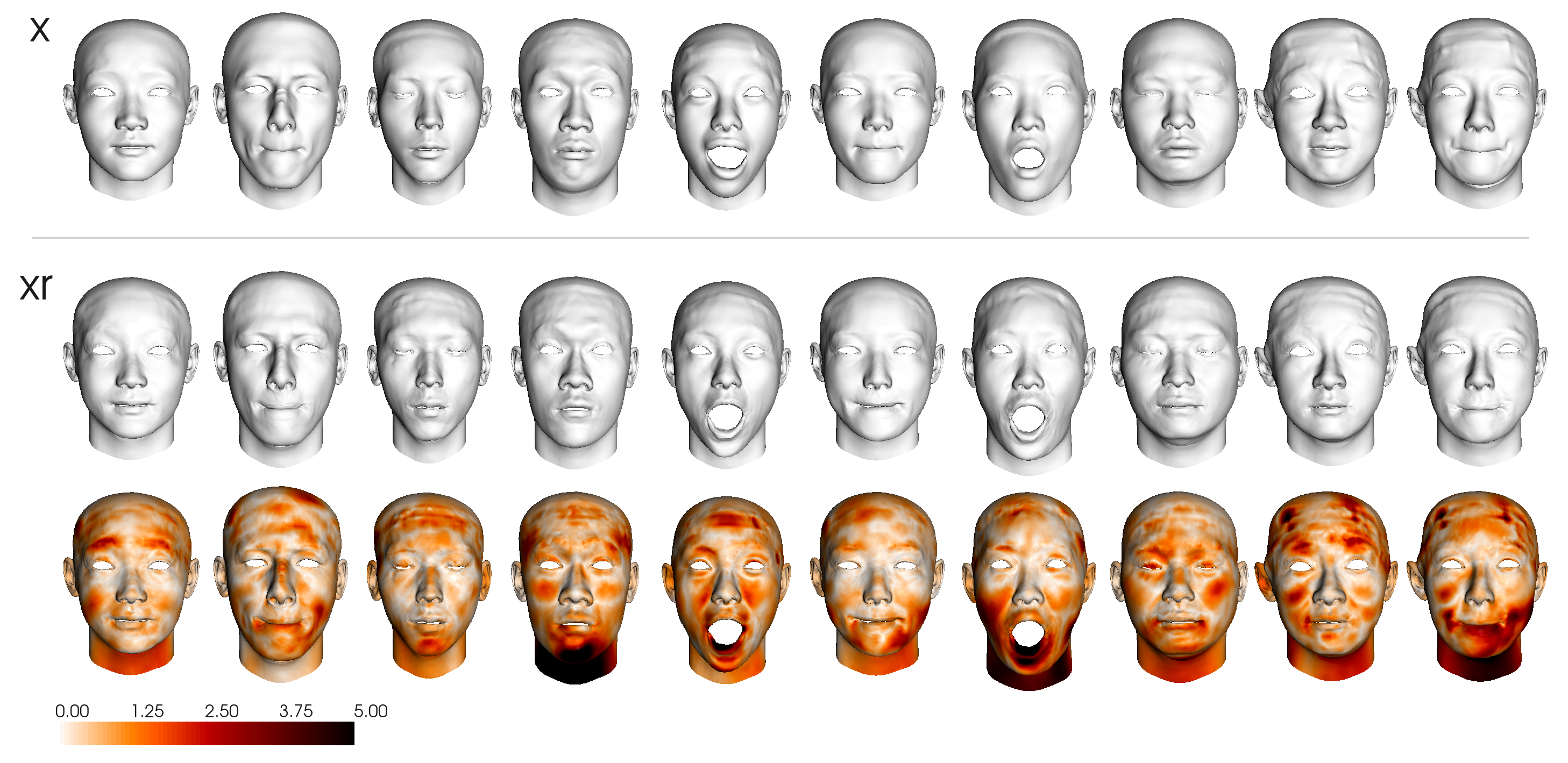}
    \caption{Reconstruction examples on FaceScape, with latent dimensions $(20, 5)$. The first row shows the input mesh. The reconstruction is displayed in the second row, while the third row shows an error map relative to the input. Values are in millimeters.}
    \label{FIG:reconstruction_facescape}
\end{figure*}

\begin{table}
    \centering
    \renewcommand{\arraystretch}{1.5}
    \begin{tabular}{CCC}
        Model & Mean & Median\\[5pt]
        \hline
        SpiralNet++ \cite{gong_spiralnet_2019} & $0.54 \pm 0.66$ & $0.32$ \\
        \hline
        FLAME \cite{li_learning_2017} & $1.45 \pm 1.65$ & $0.87$ \\
        Jiang et al.\ \cite{jiang_disentangled_2019} & $1.41 \pm 1.64$ & $1.02$ \\
        \textbf{Zhang et al. \cite{zhang_learning_2020}} & \bm{$0.67 \pm 0.75$} & \bm{$0.43$} \\
        Our method & $0.83 \pm 0.21$ & $0.77$ \\
        \hline
    \end{tabular}
    \caption{Reconstruction error on the CoMA dataset (mm). The SpiralNet++ method does not disentangle identity and expression.}
    \label{TAB:reconstruction_coma}
\end{table}

On the FaceScape dataset, we obtain a mean reconstruction error of \SI{0.81}{\mm}, with a standard deviation of \SI{0.25}{\mm} and a median of \SI{0.76}{\mm}. Metrics from other work are not available for comparison. Examples of reconstructions are given in Figure \ref{FIG:reconstruction_facescape}. We can observe that the error is spread over the face, with notable patches on the neck, eyebrows, lips and cheeks. On the \textit{mouth\_stretch} expression (columns 5 and 7), the network visibly struggles to capture the geometry and position of the lower lip.

\subsubsection{Expression Neutralization}
\label{expression_neutralization}

\begin{figure*}
    \centering
    \includegraphics[width=0.8\linewidth]{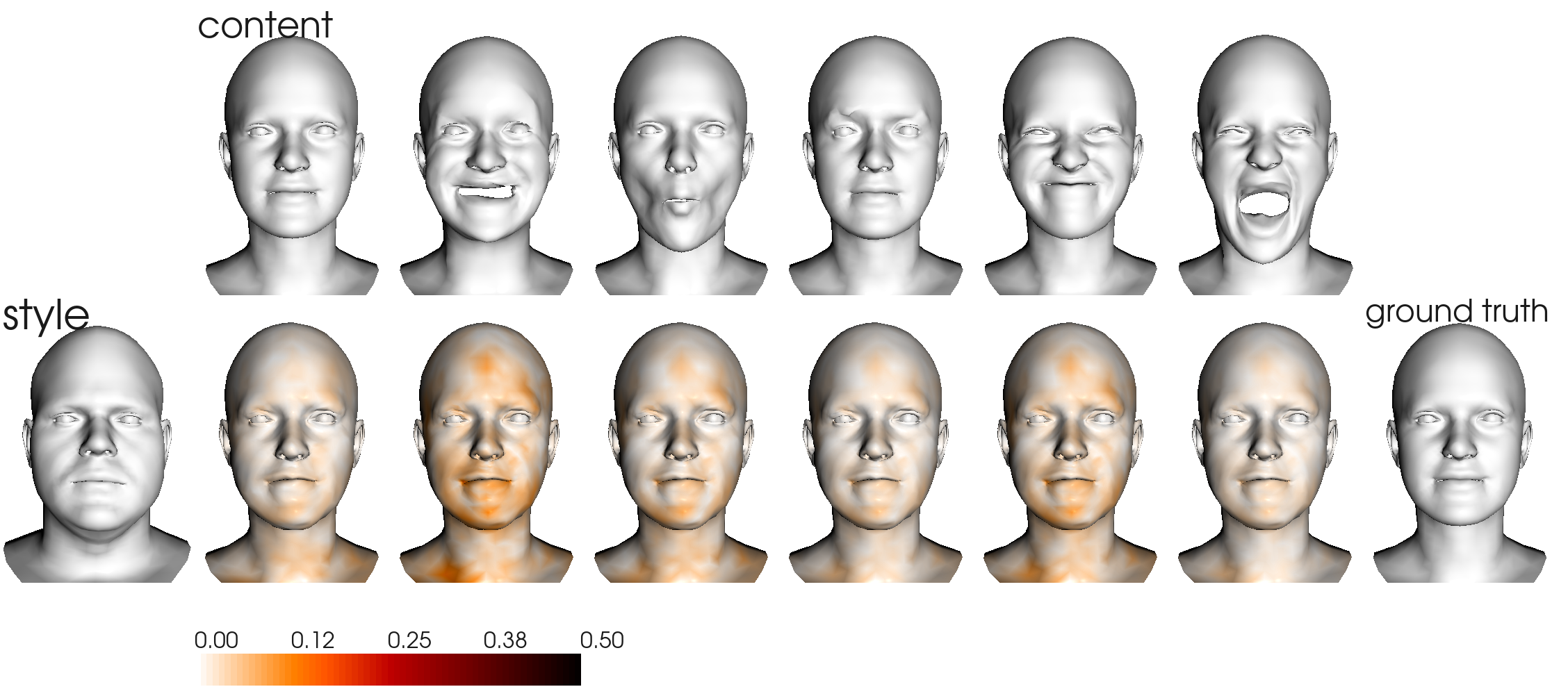}
    \caption{Variations of neutralized meshes for one subject of CoMA. The error map is relative to the mean of the outputs. Values are in millimeters. The identity decomposition error is the mean of this standard deviation computed for each subject.}
    \label{FIG:id_decomp_coma}
\end{figure*}

We evaluate our method on the expression neutralization task using two metrics. We first adapt the identity decomposition error introduced by Jiang et al.\ \cite{jiang_disentangled_2019} (and adopted by Zhang et al.\ \cite{zhang_learning_2020}) to our approach. Let $x_{c,s}$ denote a mesh of content class $c$ and style class $s$. We compute the identity decomposition as follows: for each content class $c$, we randomly select another content class $c'$ and apply its neutral style onto each $(x_{c,s})_{s \in S}$. Then, we calculate the standard deviation of the resulting meshes using AVD : $\sigma_{c} = \underset{s \in S}{\text{std}} \; Dec(E_c(x_{c,s}), E_s(x_{c',\text{neutral}}))$.
Since we are effectively applying the same neutral style on meshes $(x_{c,s})_{s \in S}$, which only differ in style, a successfully trained network is supposed to yield identical outputs. A visualization for a subject of the CoMA dataset is given in Figure~\ref{FIG:id_decomp_coma}. We report the mean of these deviations as our identity decomposition metric (Equation \eqref{eq:def_error_id_decomp}). For quantitative results, see Table \ref{TAB:id_decomp_coma}.
\begin{equation}
    E_\text{id\_decomp} = \frac{1}{|C|} \sum_{c \in C} \sigma_c  
    \label{eq:def_error_id_decomp}
\end{equation}

\begin{table}
    \centering
    \renewcommand{\arraystretch}{1.5}
    \begin{tabular}{CCC}
        Model & Mean & Median\\[5pt]
        \hline
        FLAME \cite{li_learning_2017} & $0.599$ & $0.591$  \\
        Jiang et al.\ \cite{jiang_disentangled_2019} & $0.102$ & $0.096$  \\
        Zhang et al.\ \cite{zhang_learning_2020} & $0.019$ & $0.020$ \\
        \textbf{Our method} & \bm{$0.018$} & \bm{$0.018$} \\
        \hline
    \end{tabular}
    \caption{Identity decomposition error on the CoMA dataset (mm).}
    \label{TAB:id_decomp_coma}
\end{table}

\begin{table}
    \centering
    \renewcommand{\arraystretch}{1.5}
    \begin{tabular}{CCC}
        Model & Dataset & $\text{mean} \pm \text{std} \; (\text{median})$ \\[3pt]
        \hline
        Ranjan et al.\ \cite{ranjan_generating_2018} & FaceScape & $2.88$ \\
        Kacem et al.\ \cite{kacem_disentangled_2021} & FaceScape & $2.02$ \\
        \textbf{Our method} & \textbf{FaceScape} & \bm{$1.47 \pm 1.43 \; (1.31)$} \\
        Ranjan et al.\ \cite{ranjan_generating_2018} & CoMA & $3.28$ \\
        Kacem et al.\ \cite{kacem_disentangled_2021} & CoMA & $2.73$ \\
        \textbf{Our method} & \textbf{CoMA} & \bm{$0.98 \pm 0.46 \; (0.81)$} \\
        \hline
    \end{tabular}
    \caption{Neutralization error (mm). Standard deviations and medians for other methods are not provided.}
    \label{TAB:results_neutralization}
\end{table}

In addition, we compare our results to existing methods on the FaceScape dataset by reporting the mean error between neutralized faces and the corresponding ground truth neutral. 
More precisely, we apply the following process: we randomly draw $n$ triplets $(c, c', s)_{c \in C, \; c' \in C \setminus \{c\}, \; s \in S \setminus \text{\{neutral\}}}$, i.e. two different content classes and a non-neutral style class. For each triplet $(c,c',s)$, we apply the style of $x_{c', \text{neutral}}$ onto $x_{c, s}$ and compare the output with ground truth $x_{c, \text{neutral}}$ (Equation \eqref{eq:def_error_neutralization}). Results are presented in Table \ref{TAB:results_neutralization}.
\begin{equation}
    E_\text{neu} = \frac{1}{n} \sum_{(c,c',s)} \text{AVD}(x_{c, \text{neutral}}, Dec(E_c(x_{c, s}), E_s(x_{c', \text{neutral}})))  
    \label{eq:def_error_neutralization}
\end{equation}

Note that our model is trained in the unpaired setting. We do not explicitly pair expressive faces with their neutral counterpart, contrary to the compared methods. These metrics allow us to evaluate our method against the state of the art for expression neutralization. However, this task is only one particular case of our network’s capabilities.

\subsubsection{Expression Transfer}
\label{exptransfer}

We now evaluate our model on the more general expression transfer task. We introduce a metric similar to the neutralization error above and provide a baseline. We randomly draw $n$ triplets exactly as described above: each triplet $(c,c',s)$ contains two different content classes and a non-neutral style class. This time, the style of $x_{c', s}$ is applied onto $x_{c, \text{neutral}}$: $x_t = Dec(E_c(x_{c, \text{neutral}}), E_s(x_{c', s}))$. The output $x_t$ is compared to the ground truth $x_{c, s}$.
\begin{equation}
    E_\text{transfer} = \frac{1}{n} \sum_{(c,c',s)} \text{AVD}(x_{c, s}, x_t)  
    \label{eq:def_error_transfer}
\end{equation}

\begin{figure*}
    \centering
    \includegraphics[width=\linewidth]{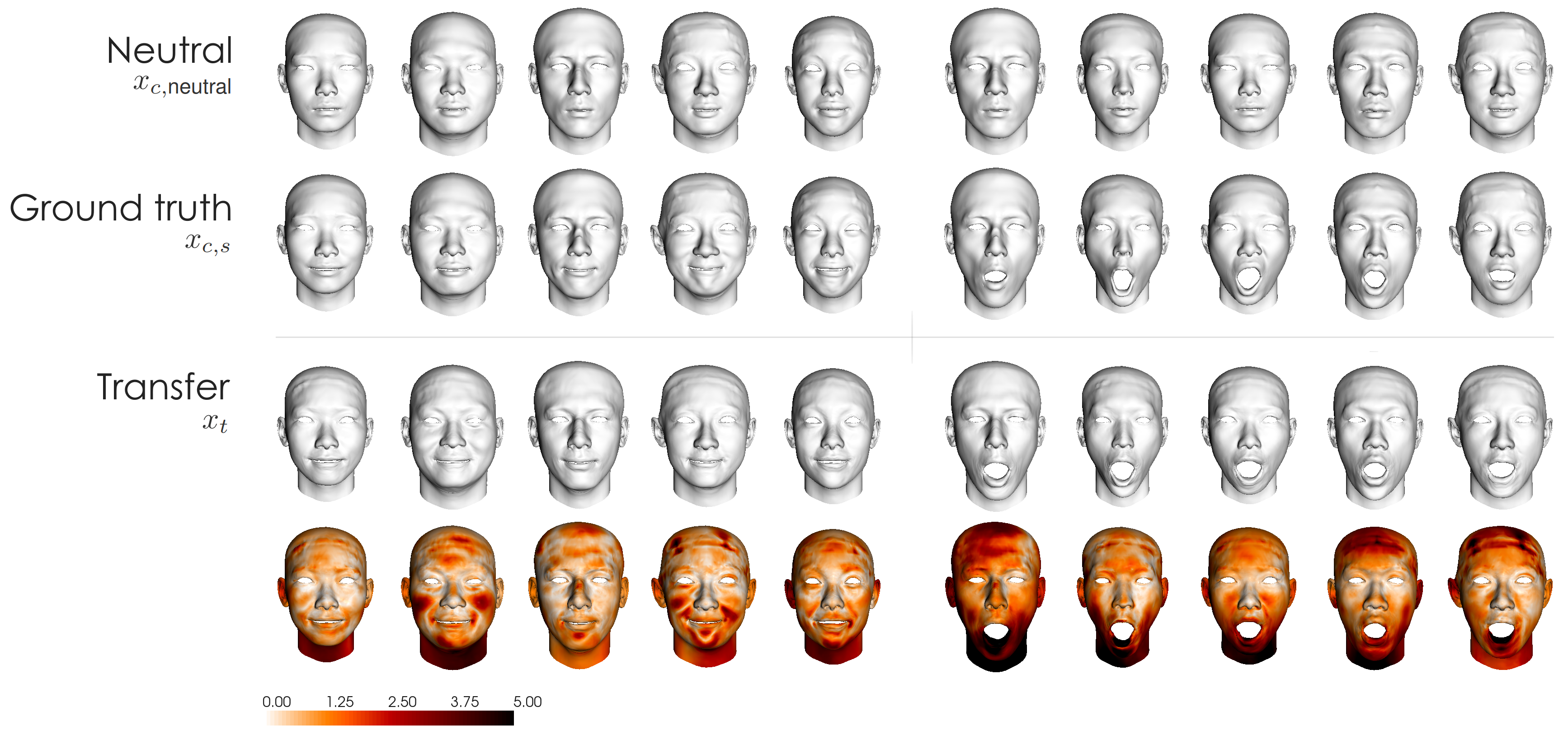}
    \caption{Expression transfer results for two expressions (smile on the left, mouth\_stretch on the right). The transfer outputs are computed as described in Section \ref{exptransfer}.}
    \label{FIG:exp_transfer_facescape}
\end{figure*}

In Table \ref{TAB:results_exptransfer}, we report our expression transfer error on the FaceScape dataset for each style class $s$. It can be observed that some expressions are associated with much higher transfer errors than others. The most difficult expressions seem to be the ones farthest apart from neutral. Figure \ref{FIG:exp_transfer_facescape} shows visual examples for \textit{smile} and \textit{mouth\_stretch}, the expressions with the lowest and highest error. It illustrates one of the challenges of the task particularly well. The first subject in the \textit{mouth\_stretch} expression opens his jaw less than most subjects in the dataset. In the transfer output, the lower part of the face is lower than it should be, because the corresponding $x_{c',s}$ (not shown) whose style is transferred has a more open jaw. During training, this would penalize the network. However, this difference could perhaps be attributed to a different interpretation of the expression by the subject (i.e. for the same expression, two scans can be different due to factors other than identity).

\begin{table}
    \centering
    \begin{tabular}{RCC}
        Expression & Our Method & Bs transfer \\[3pt]
        \hline
        smile & \textbf{1.23} & 2.11\\
        anger & \textbf{1.32} & 3.25\\
        sadness & \textbf{1.45} & 4.10\\
        grin & \textbf{1.57} & 2.60\\
        mouth\_stretch & \textbf{2.02} & 3.26\\
        mouth\_left & \textbf{1.61} & 2.78\\
        mouth\_right & \textbf{1.54} & 2.45\\
        dimpler & \textbf{1.42} & 2.20\\
        jaw\_left & \textbf{1.53} & 2.24\\
        jaw\_right & \textbf{1.63} & 2.59\\
        jaw\_forward & \textbf{1.62} & 2.43 \\
        chin\_raiser & \textbf{1.65} & 2.82\\
        lip\_puckerer & \textbf{1.56} & 2.36\\
        lip\_funneler & \textbf{1.80} & 2.64\\
        lip\_roll & \textbf{1.52} & 2.50\\
        cheek\_blowing & \textbf{1.77} & 3.55\\
        eye\_closed & \textbf{1.23} & 1.86\\
        brow\_raiser & \textbf{1.30} & 3.06\\
        brow\_lower & \textbf{1.41} & 3.27\\
        \textit{all} & \textbf{1.54} &2.74 \\
        \hline
    \end{tabular}
    \caption{Expression transfer errors on FaceScape (mm) with our method and the blendshape transfer.}
    \label{TAB:results_exptransfer}
\end{table}

\begin{figure}
    \centering
    \includegraphics[width=\linewidth]{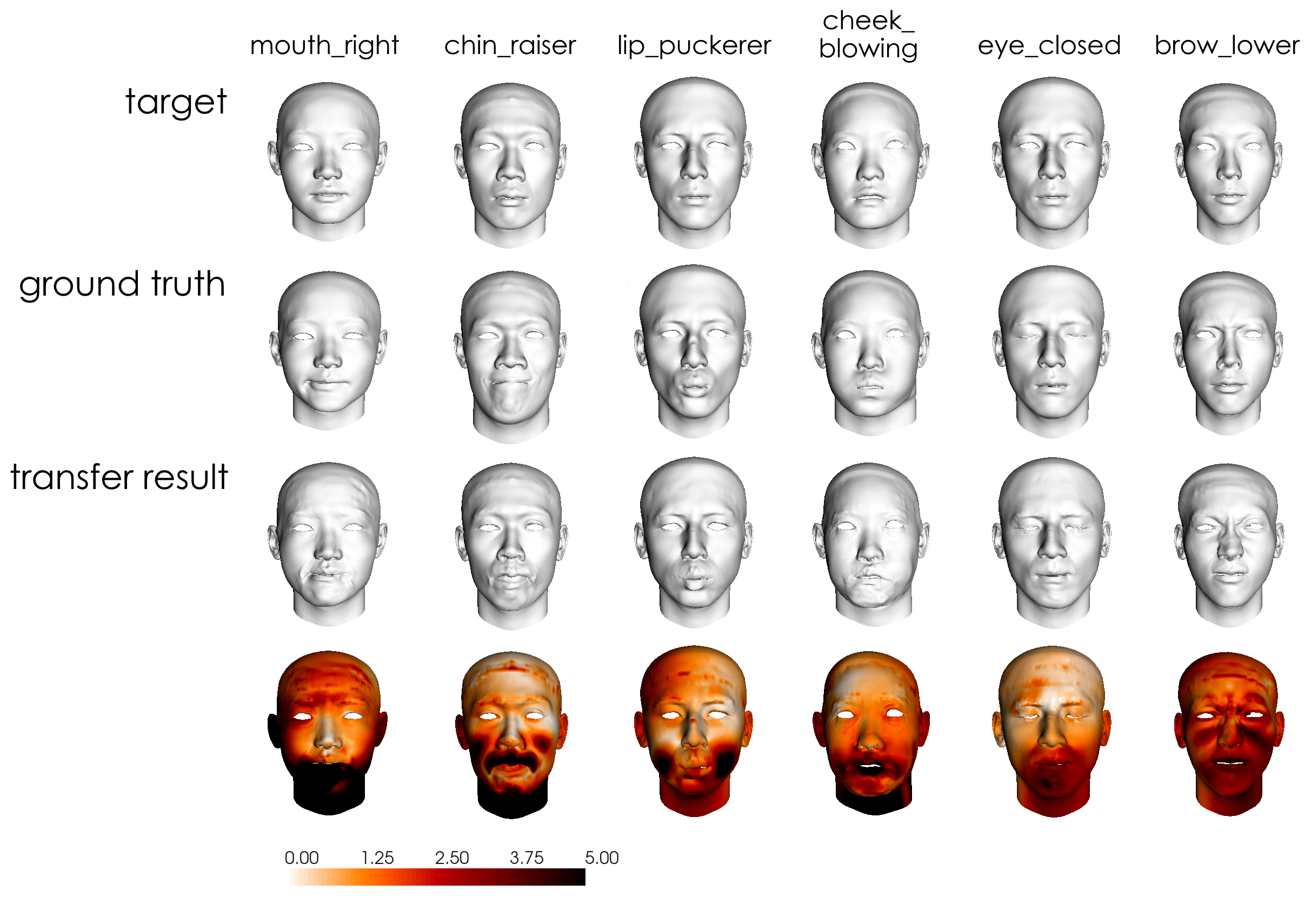}
    \caption{Blendshape transfer results. Expressions are tranfered from a source (not shown) to the target}
    \label{FIG:blendshape_transfer_results}
\end{figure}

Finally we compare our method to traditional blendshape transfer \cite{sumner2004deformation}. Results are depicted in Figure~\ref{FIG:blendshape_transfer_results}. A visual comparison to our method is shown on Figure \ref{FIG:blendshape_transfer_comparison}. For the two selected expressions, the blendshape transfer method applies the expression geometry to the target as it is on the source face, without taking into account the target morphology. Our method leads to more realistic outputs that look closer to the ground truth. Objective measurements shows that our method has smaller errors, as reported in Table~\ref{TAB:results_exptransfer}.

\section{Limitations and Future Work}
\label{limitations-further-work}

We have developed a network architecture that adapts style transfer techniques to 3D faces and demonstrated results that are competitive with the state of the art reconstruction, neutralization and expression transfer.
However, there are several ways in which this work could be extended.

\textbf{Interpolation} : we visually investigated our model's ability to interpolate between two faces. Leveraging our low-dimensional representations, we perform linear interpolations in the content and style spaces. Let $A$ and $B$ denote two style codes obtained by encoding two scans of the same subject in a different facial expression. To interpolate, we move along the style vector $\overrightarrow{AB}$ in fixed increments of $s \times ||\overrightarrow{AB}||$ in the style space. Similarly, we can interpolate in content space. Figure \ref{FIG:interp_exp_facescape_full} shows results of interpolating between expressions of the same subject on the FaceScape dataset. In Figure \ref{FIG:interp_id_facescape}, we interpolate between different subjects in their neutral expression.

Geometric approaches (e.g. blendshapes) have explicit control over the intensity of the expression being applied. On the contrary, our method in its current state does not have any additional constraint on the latent spaces. Hence the results of these interpolations are completely dependent on the structure of the data on which the network was trained. We find that the network learns a much more regular representation for the content space than the style space. We attribute this to the large amount of subjects in the FaceScape dataset, in comparison to the more limited variation of the 20 expressions.

\textbf{Extrapolation} : the previous interpolation method was extended to investigate the network's ability to extrapolate from the training distribution. In Figure \ref{FIG:extrap_neutral_exp_facescape}, we extrapolate in style space along a vector going from the neutral code of a subject to its style code for another expression. We find that while the network can sometimes succeed in amplifying the expression (e.g. \textit{mouth\_stretch} and \textit{cheek\_blowing} on Figure \ref{FIG:extrap_neutral_exp_facescape}), it is also susceptible to changing some features in unexpected ways (e.g. the \textit{dimpler} expression on Figure \ref{FIG:extrap_neutral_exp_facescape}).

Figure \ref{FIG:extrap_content_facescape} shows extrapolation in the content space. Our method is able to amplify the differences between two faces. Similarly to our findings for interpolation, we can observe that the extrapolations in content space are less prone to adding noise and warping unwanted features than those in the style space. These interpolation and extrapolation experiments are only preliminary tests. Due to the subjective nature of this task, a user study would need to be conducted to validate these results.

\textbf{Increased realism} : while we currently only predict geometry, one could use the data available in FaceScape to learn prediction of displacement maps and textures from the latent codes and generate much more realistic faces. Our network’s ability to decouple identity and expression factors could be well-suited to modelling the subtle expression-specific texture variations, a capability demonstrated by Chandran et al.\ \cite{chandran_semantic_2020}.

\textbf{Investigate bias} : it is common for deep learning methods to display biases in the output distribution. While the gender of the subjects in the FaceScape dataset is distributed quite evenly, there is a clear bias toward 18 to 24 year-old Asians. It would be beneficial to study how this bias is reflected in the outputs and whether the network is able to generalize to more diverse faces.

\textbf{Generative model} : currently, sampling random content or style codes is very unlikely to provide satisfying results. In order to add this generative ability to the network, our autoencoder could be turned into a Variational Autoencoder~\cite{kingma_auto-encoding_2014} by adding a regularization term (e.g. Kullback-Leibler divergence loss) and encoding distributions instead of single points. This could also improve our ability to interpolate in latent space.

\textbf{Unseen expressions} : we could take a step further toward adapting the performance of image translation networks to the 3D domain by adding few-shot inference for unseen style classes. Given a few samples of a new style class, the model would be able to translate meshes to that new class. This capability is demonstrated in 2D by the FUNIT model~\cite{liu_few-shot_2019}. 

\textbf{Application to other domains} : finally, while our approach was only tested on face datasets, applying it to data from other domains could yield interesting results. The image-to-image translation models from which it is inspired have been able to operate on animals, shapes of shoes and handbags, seasonal variations of landscapes, painting styles, etc. The challenge here will be the availability of large, labeled 3D datasets.

\section{Conclusion}
\label{conclusion}

We have presented FaceTuneGAN, a novel 3D face model representation decomposing and encoding separately facial identity and facial expression. The proposed method is a first adaptation to the 3D domain of existing work for transferring features of images from one style class to another. Adapting several existing techniques to the 3D domain, this new style-based adversarial autoencoder architecture can capture identity and expression features of the input 3D face in separated low-dimensional spaces. Two encoders respectively extract the content (identity) and style (expression) information, which the decoder takes as inputs to reconstruct a 3D face. Additionally, a discriminator is used in an adversarial training scheme to regularize the training, enforcing the output to be realistic and of the correct style class. This method is shown to be better with previous state-of-the-art approaches on several tasks, outperforming them on several metrics, such as identity decomposition and neutralisation. 

This method could bolster the creation of digital characters, allowing to accurately transfer expressions between various facial morphologies. The architecture of the method containing nothing specific to expressions, it also has the potential to be adapted to applied to other style transfer applications for 3D models.

%

%

%

%

\section{Acknowledgments}
This project has received funding from the Association Nationale de la Recherche et de la Technologie under CIFRE agreement No 2018/1656, and from the European Union's Horizon 2020 research and innovation program under grant agreement No 952147.

\bibliographystyle{cas-model2-names}
\bibliography{main}

\appendix
\section{Architecture Details}
\label{appendix:archi-details}

\begin{figure}
    \centering
    \includegraphics[width=\linewidth]{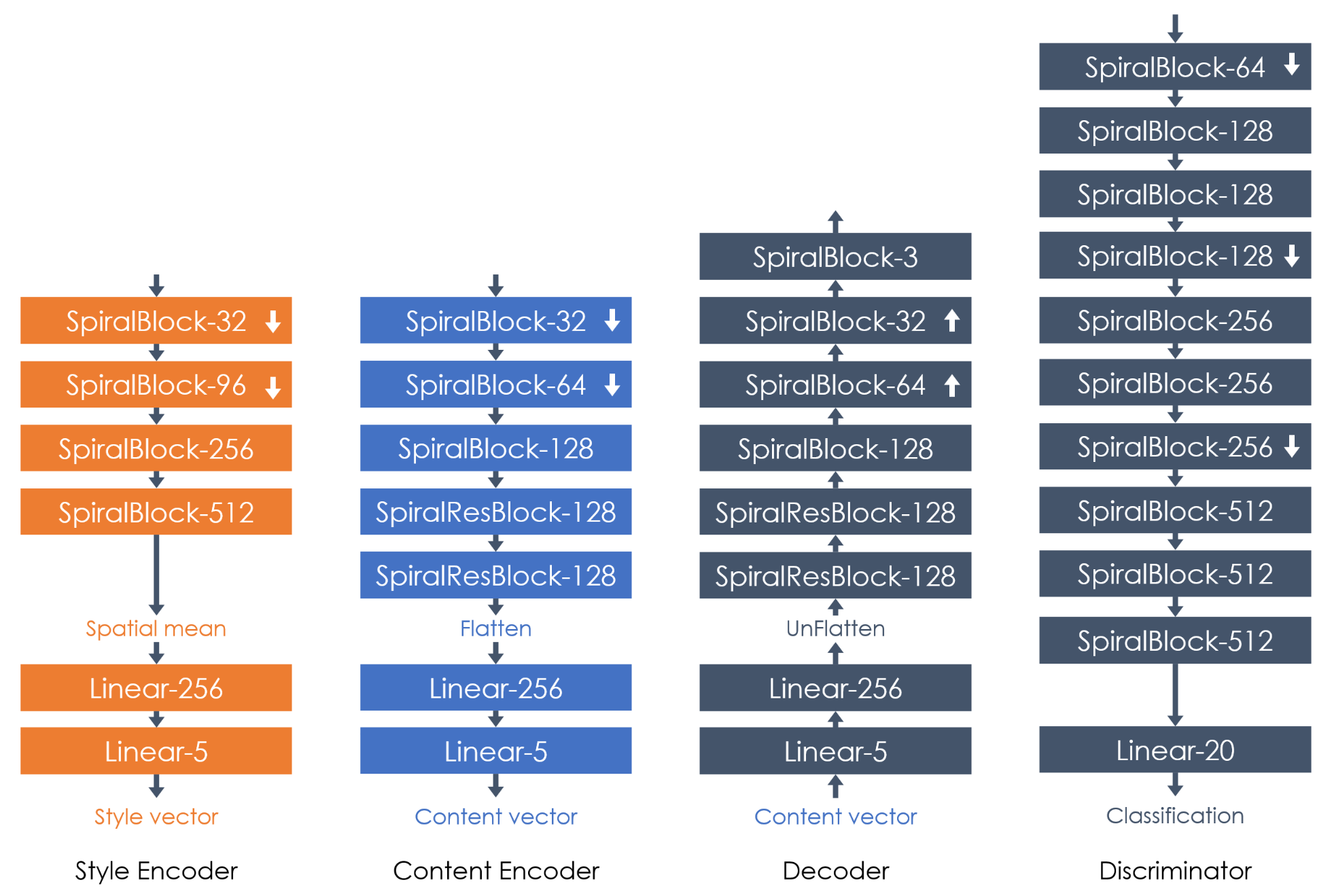}
    \caption{Composition of each module for training on FaceScape. The down/up arrows respectively indicate down-sampling and up-sampling layers, always by a factor of 4.}
    \label{FIG:archi_specifics_facescape}
\end{figure}

\begin{figure}
    \centering
    \includegraphics[width=\linewidth]{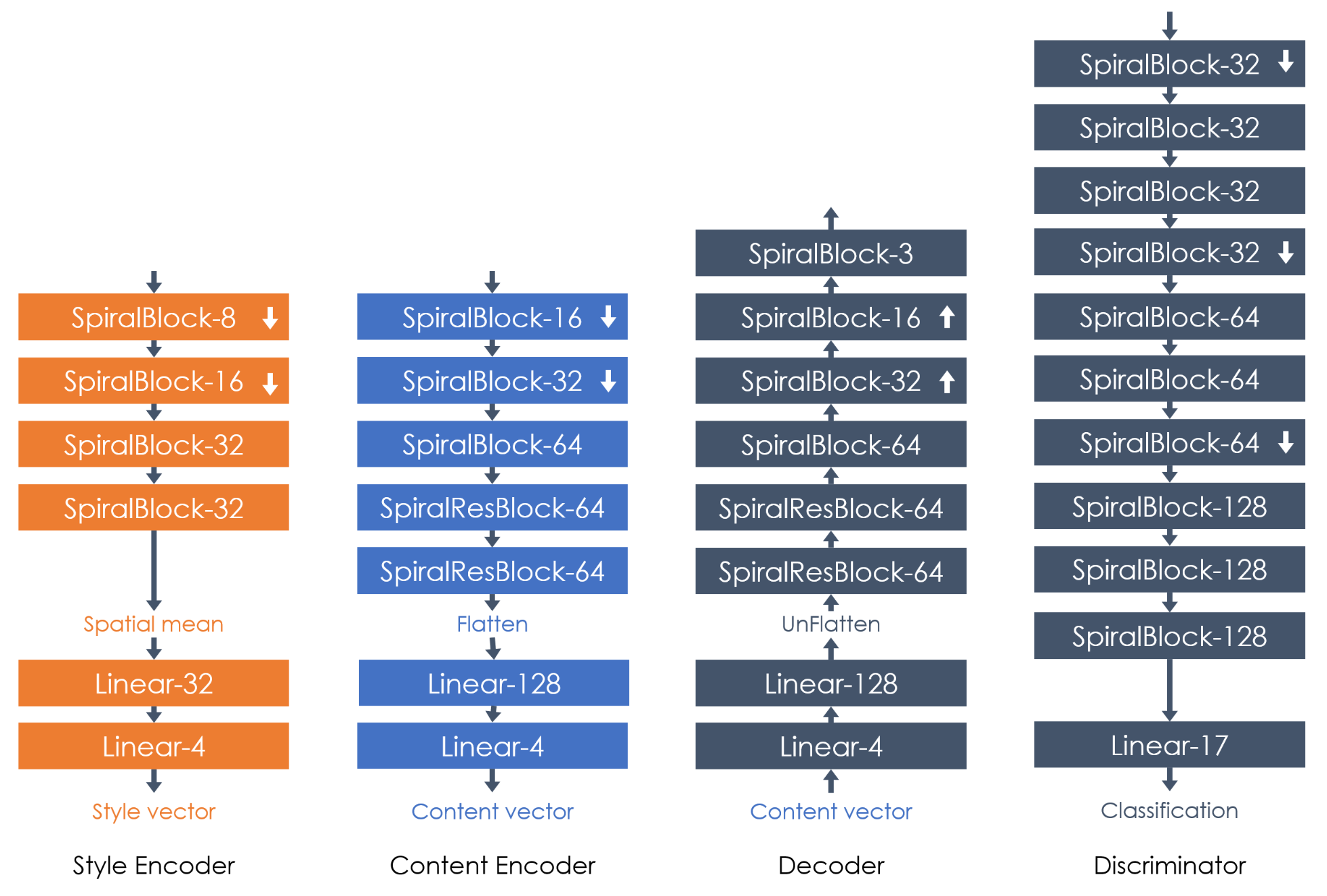}
    \caption{Network modules for training on CoMA. The down/up arrows respectively indicate down-sampling and up-sampling layers, always by a factor of 4.}
    \label{FIG:archi_specifics_coma}
\end{figure}

This section details the specific layers of our network. Due to the different amount of vertices in the datasets (27k in FaceScape, 5k in CoMA), the architecture is tweaked differently for each. Figure \ref{FIG:archi_specifics_facescape} and \ref{FIG:archi_specifics_coma} respectively depict the composition of our network modules for training on CoMA and FaceScape. All linear layers use ReLU activations, except for the final discriminator layer which uses a Softmax activation. All SpiralBlocks and SpiralResBlocks use Exponential Linear Unit (ELU) activations \cite{clevert_fast_2016}. The style encoder and discriminator use no normalization. The content encoder uses Instance Normalization layers on all SpiralBlocks and SpiralResBlocks. The decoder uses Adaptive Instance Normalization on its two first SpiralResBlocks and first SpiralBlock to inject the style information.

The mapping function $M$ that converts the style codes into parameters for the AdaIN normalization layers of the decoder is composed of fully-connected layers. No normalization is used, and all layers use ReLU activations except for the last one. On CoMA, we use the following configuration: $4 \rightarrow 128 \rightarrow 128 \rightarrow 16 \rightarrow N_{\text{AdaIN}}$ where $N_{\text{AdaIN}}$ is the number of AdaIN parameters in the decoder. On FaceScape, we use more layers and a larger size: $5 \rightarrow 256 \rightarrow 256 \rightarrow 256 \rightarrow 256 \rightarrow 256 \rightarrow 256 \rightarrow 16 \rightarrow N_{\text{AdaIN}}$.

\section{CoMA dataset}

\begin{figure}
    \centering
    \captionsetup{justification=centering}
    \includegraphics[width=\linewidth]{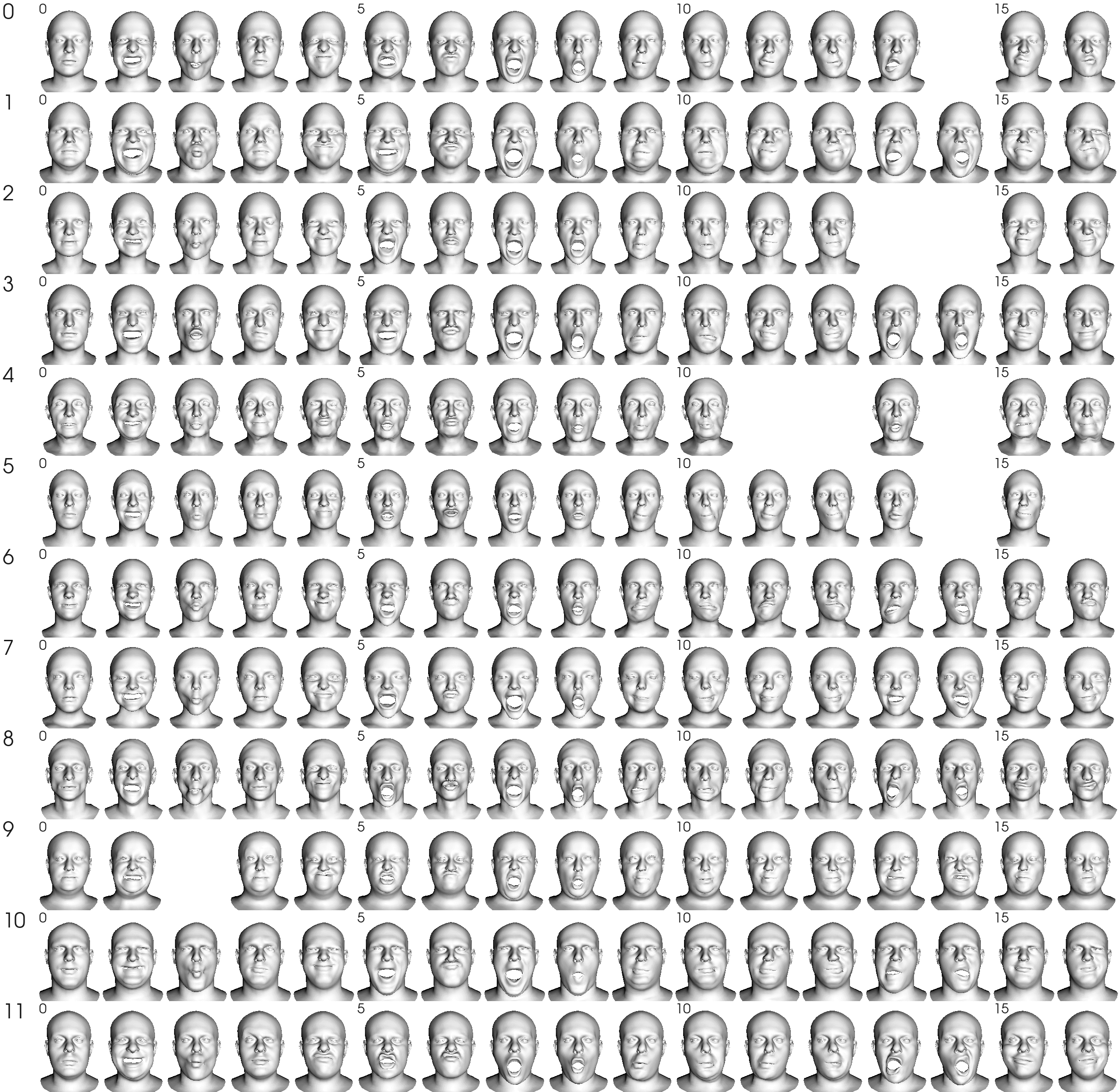}
    \caption{Selected frames on CoMA. Entries are missing when the expression performed by the subject varies significantly from the rest.}
    \label{FIG:coma}
\end{figure}

Figure \ref{FIG:coma} shows one selected frame for each (subject, expression) pair of the CoMA \cite{ranjan_generating_2018} dataset.  

\section{Interpolations}

\begin{figure}
    \centering
    \includegraphics[width=0.6\linewidth]{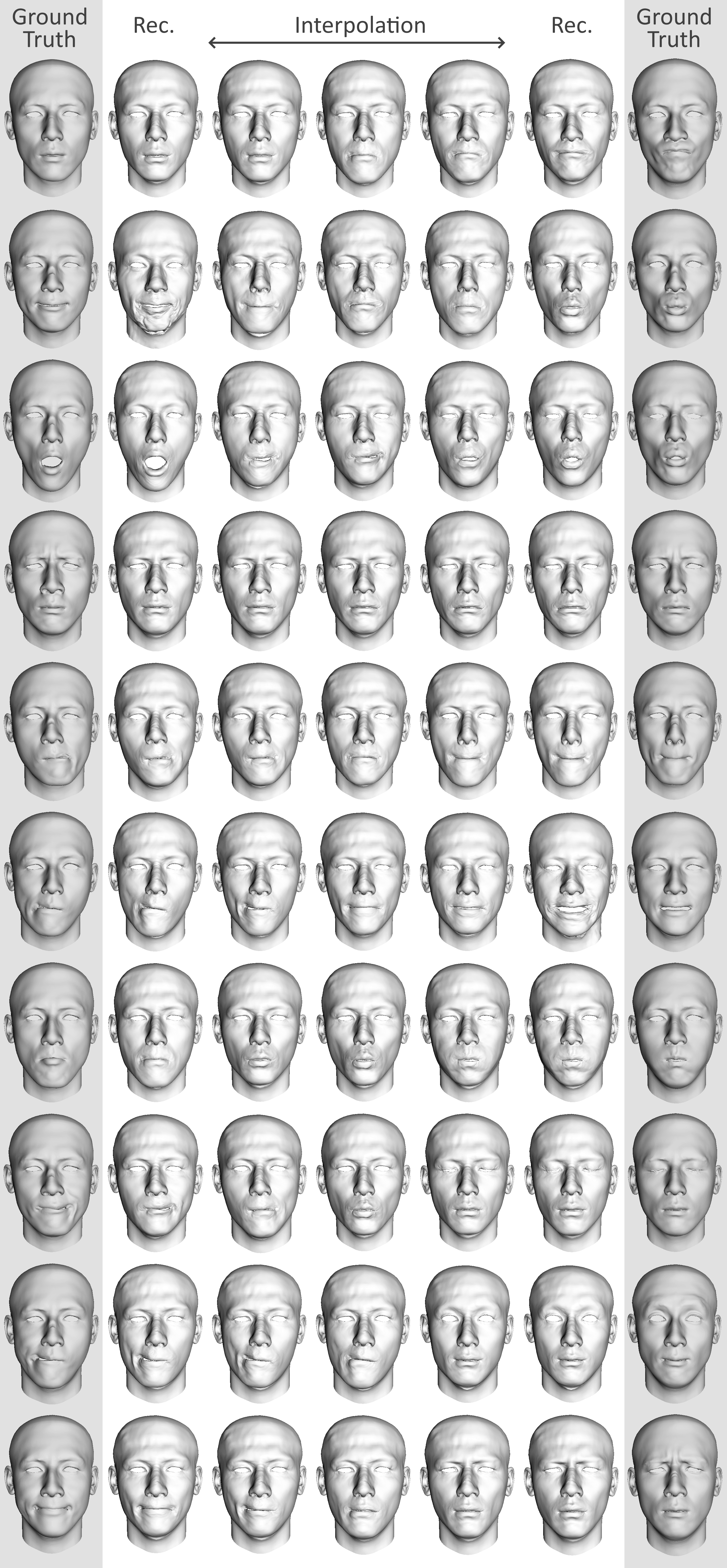}
    \caption{Expression interpolation results on FaceScape (style space). The ground truth scans are provided on the leftmost and rightmost columns. The second and second-to-last columns show their reconstruction. In the middle three columns, we interpolate along the style vector ($s = 0.25$).}
    \label{FIG:interp_exp_facescape_full}
\end{figure}

\begin{figure}
    \centering
    \includegraphics[width=\linewidth]{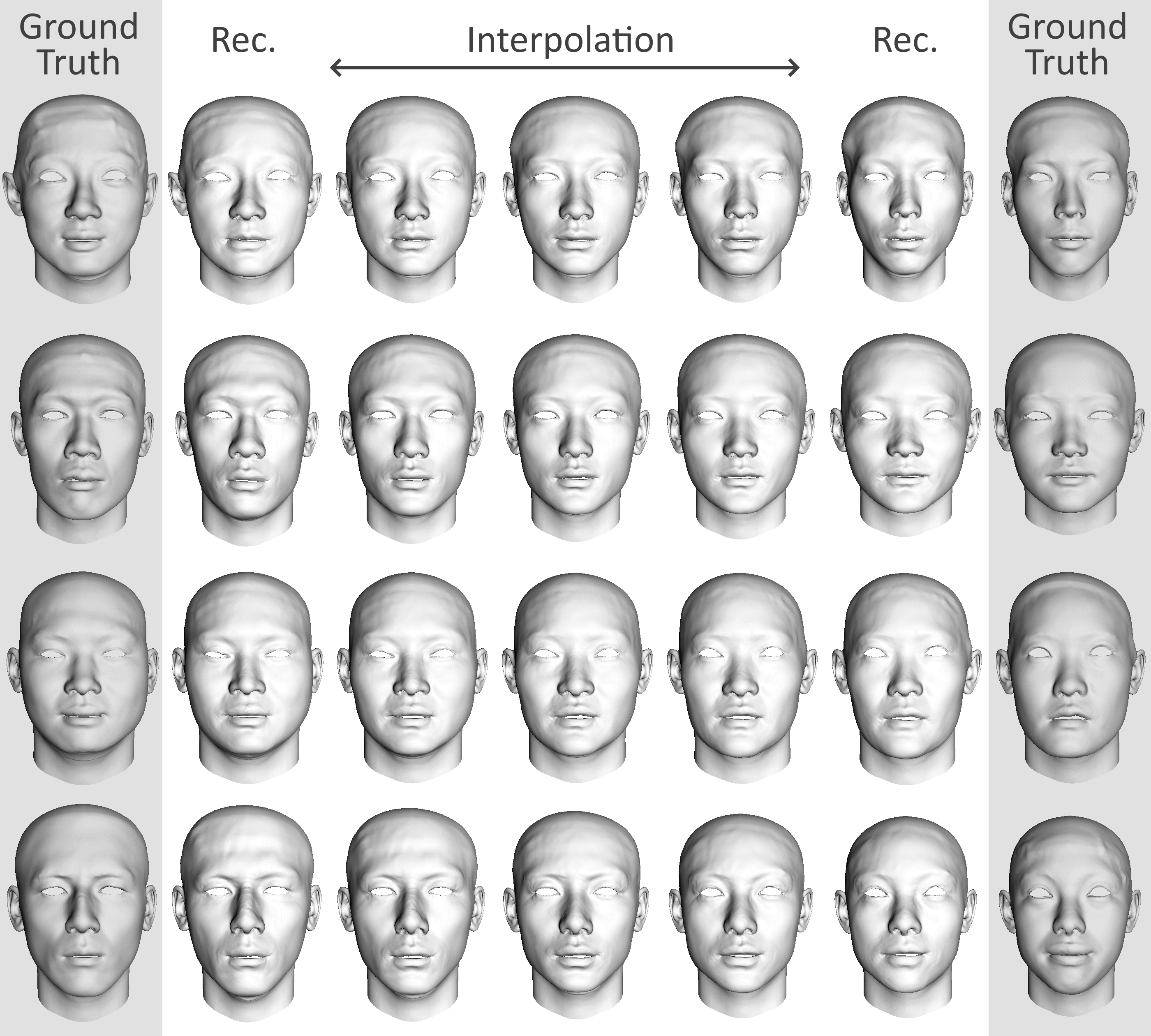}
    \caption{Identity interpolation results on FaceScape (content space). The ground truth scans are provided on the leftmost and rightmost columns. The second and second-to-last columns show their reconstruction. In the middle three columns, we interpolate along the content vector ($s = 0.25$).}
    \label{FIG:interp_id_facescape}
\end{figure}

Expressions and identities interpolations are provided in Figures \ref{FIG:interp_exp_facescape_full} and \ref{FIG:interp_id_facescape} respectively.

\section{Expression Extrapolations}

\begin{figure}
    \centering
    \includegraphics[width=\linewidth]{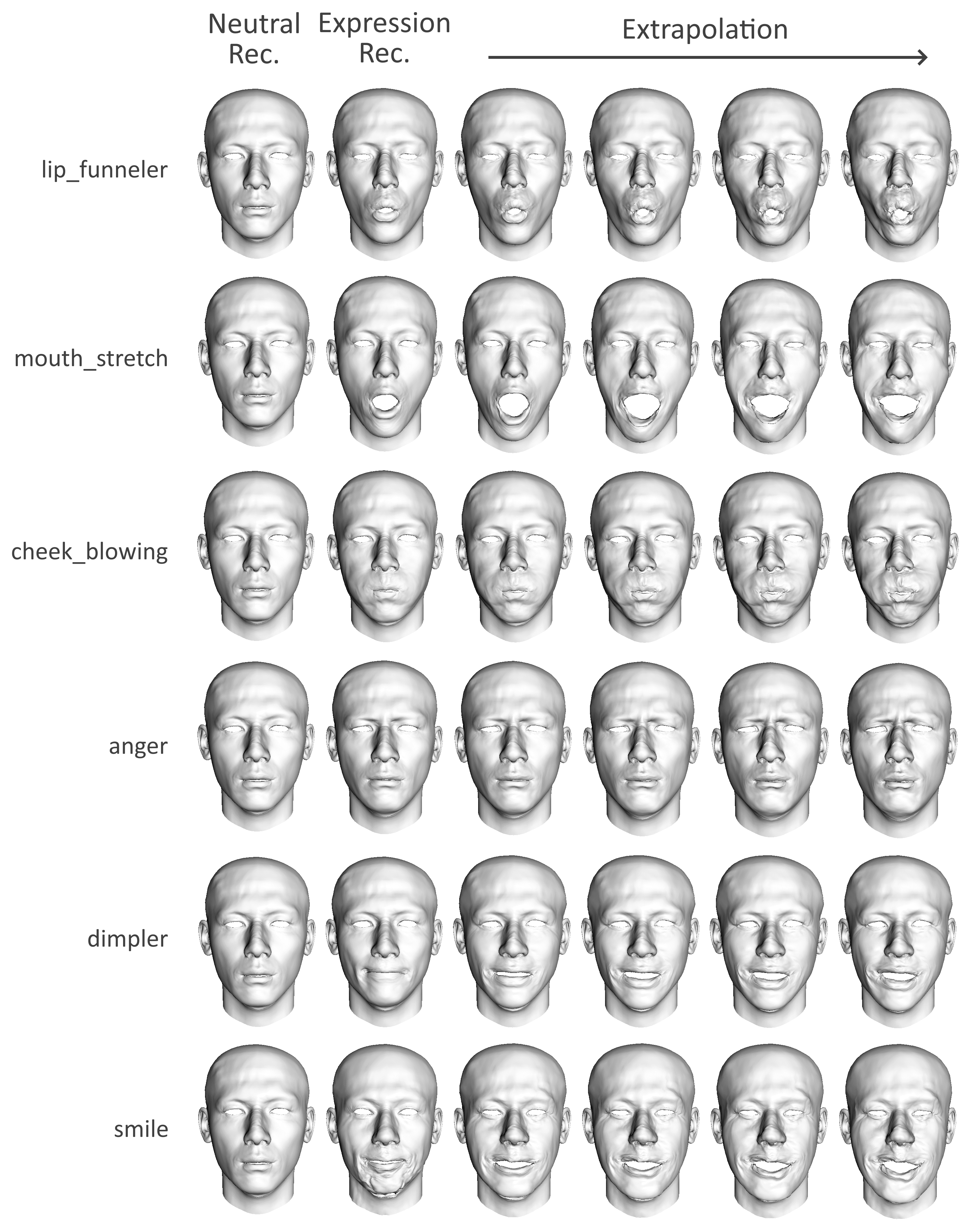}
    \caption{Extrapolation in style space. The two leftmost columns are reconstructions of the neutral and expression scans. We gradually extrapolate along the style vector ($s = 0.5$).}
    \label{FIG:extrap_neutral_exp_facescape}
\end{figure}

\begin{figure*}
    \centering
    \includegraphics[width=1.0\linewidth]{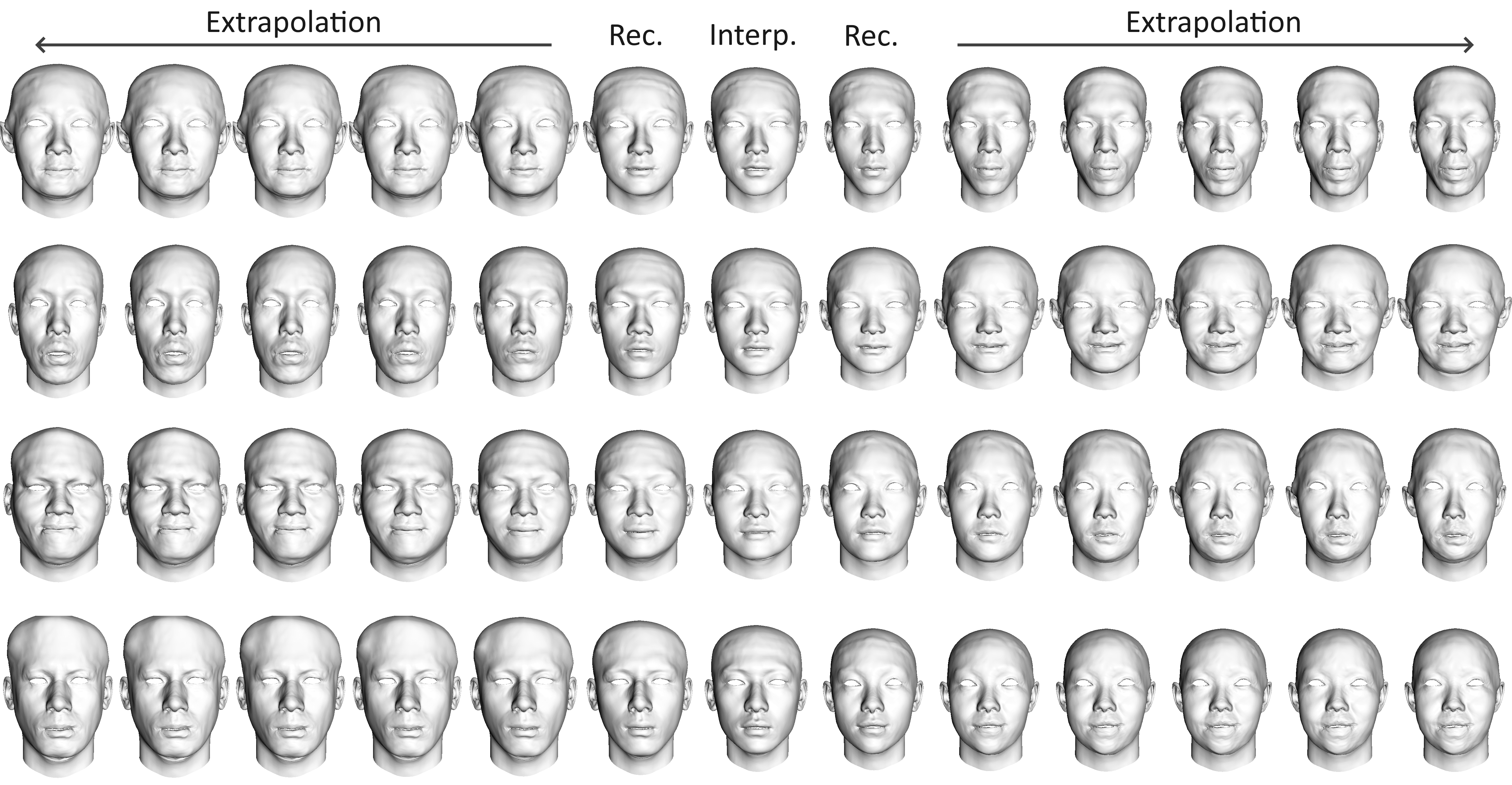}
    \caption{Extrapolation in content space. The reconstructions of two neutral scans are shown in the columns marked "Rec.". We move along their relative content vector from the middle to the outer columns ($s = 0.5$).}
    \label{FIG:extrap_content_facescape}
\end{figure*}

Expressions and identities extrapolations are provided in Figures \ref{FIG:extrap_neutral_exp_facescape} and \ref{FIG:extrap_content_facescape} respectively.

\end{document}